# A FEATURE EXTRACTION TECHNIQUE BASED ON CHARACTER GEOMETRY FOR CHARACTER RECOGNITION

Dinesh Dileep

*Abstract*—This paper describes a geometry based technique for feature extraction applicable to segmentation-based word recognition systems. The proposed system extracts the geometric features of the character contour. This features are based on the basic line types that forms the character skeleton. The system gives a feature vector as its output. The feature vectors so generated from a training set, were then used to train a pattern recognition engine based on Neural Networks so that the system can be benchmarked.

## I. INTRODUCTION

THIS paper has been inspired from the work in [1]. he literature explains many high accuracy recognition systems for seperated handwritten numerals and characters. However feature extraction based on local and global geometric features of the character skeleton has not been investigated much. The algorithm proposed concentrates on the same. It extracts different line types that forms a particular character. It also concentrates on the positional features of the same. The feature extraction technique explained was tested using a Neural Network which was trained with the feature vectors obtained from the system proposed.

## II. OVERVIEW

The paper starts with methods in Image preprocessing. Section 3 gives steps involved in image preprocessing. The sections that follow explain the algorithm. Finally section 10 concludes the discussion.

## III. IMAGE PREPROCESSING

Image preprocessing involves the following steps
1) Character Extraction from Scanned Document.
2) Binarization.
3) Background Noise removal.
4) Skeletonization.

The paper assumes that the input Image is available after undergoing all this processes. Some excellent papers on this steps are given in the references.

Dinesh Dileep is with the Department of Electronics and Communication Engineering, Amrita School of Engineering, Kollam, Kerala, INDIA pin-691571 e-mail:dineshdileep@gmail.com

## IV. UNIVERSE OF DISCOURSE

Universe of discourse is defined as the shortest matrix that fits the entire character skeleton. The Universe of discourse is selected because the features extracted from the character image include the positions of different line segments in the character image. So every character image should be independent of its Image size.

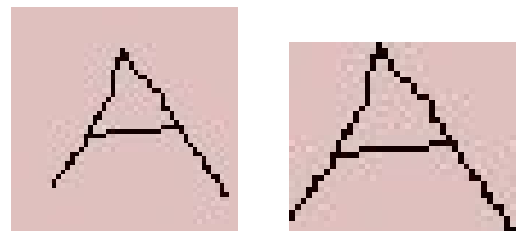

(a) Original Image    (b) Universe of Discourse

## V. ZONING

After the universe of discourse is selected, the image is divided into windows of equal size, and the feature is done on individual windows. For the system implemented, two types of zoning were used. The image was zoned into 9 equal sized windows. Feature extraction was applied to individual zones rather than the whole image. This gives more information about fine details of character skeleton. Also positions of different line segments in a character skeleton becomes a feature if zoning is used. This is because, a particular line segment of a character occurs in a particular zone in almost cases. For instance, the horizontal line segment in character 'A' almost occurs in the central zone of the entire character zone.

## VI. STARTERS, INTERSECTIONS AND MINOR STARTERS.

To extract different line segments in a particular zone, the entire skeleton in that zone should be traversed. For this purpose, certain pixels in the character skeleton were defined as starters, intersections and minor starters.

### A. Starters

Starters are those pixels with one neighbour [1] in the character skeleton.Before character traversal starts, all the starters in the the particular zone is found and is populated in a list.

---

[1] Neighbourhood includes all the pixels that immediately surrounds the pixel under consideration.

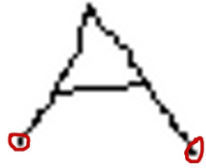

Fig. 1. Starters are rounded

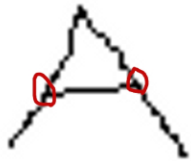

Fig. 2. Intersections

## B. Intersections

The definition for intersections is some what more complicated. The necessary but insufficient criterion for a pixel to be a intersection is that it should have more than one neighbours. A new property called true neighbours is defined for each pixel. Based on the number of true neighbours for a particular pixel, it is classified as an intersection or not. For this, neighbouring pixels are classified into two categorires, direct pixels and diagonal pixels. Direct pixels are all those pixels in the neighberhood of the pixel under consideration in the horizontal and vertical directions. Diagonal pixels are the remaining pixels in the neighberhood which are in a diagonal direction to the pixel under consideration. Now for finding number of true neighbours for the pixel under consideration, it has to be classified further based on the number of neighbours it have in the character skeleton.

Pixels under consideration are classifed as those with

- 3 neighbours:If any one of the direct pixels is adjacent to anyone of the diagonal pixels, then the pixel under consideration cannot be an intersection, else if if none of the neighbouring pixels are adjacent to each other then its a intersection.
- 4 neighbours:If each and every direct pixel have a adjacent diagonal pixel or vice-versa, then the pixel under consideration cannot be considered as a intersection.
- 5 or neighbours:If the pixel under consideration have five or more neighbours, then it is always considered as a intersection

Once all the intersections are identified in the image, then they are populated in a list.

## C. Minor starters

Minor starters are found along the course of traversal along the character skeleton. They are created when pixel under consideration have more than two neighbours. There are two conditions that can occur

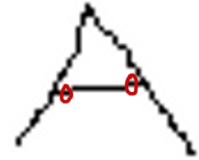

Fig. 3. Minor starters

- Intersections: When the current pixel is an intersection. The current line segment will end there and all the unvisited neighbours are populated in the minor starters list.
- Non-intersections: Situations can occur where the pixel under consideration has more than two neighbours but still its not a intersection. In such cases, the current direction of traversal is found by using the position of the previous pixel. If any of the unvisited pixels in the neighberhood is in this direction, then it is considered as the next pixel and all other pixels are populated in the minorstartes list.If none of the pixels is not in the current direction of travesal, then the current segment is ended there and all the pixels in the neighberhood are populated in the minor starters list.

When the algorithm proposed is applied to character 'A', in most cases , the minor starters found are given in the image.

## VII. CHARACTER TRAVERSAL

Character traversal starts after zoning is done on the image. Each zone is individually subjected to the process of extracting line segments. For this purpose, first the starters and intersections in the zone are found and then populated in a list. Minor starters are found along the course of traversal. Algorithm starts by considering the starters list. Once all the starters are processed, the minor starters obtained so are processed. After that, the algorithm starts with the minor starters. All the line segments obtained during this process are stored, with the positions of pixels in each line segment. Once all the pixels in the image are visited, the algorithm stops.

### A. An Example

The following example is to illustrate the algorithm explained earlier.

| 1 | 0 | 0 | 0 | 1 |
|---|---|---|---|---|
| 0 | 1 | 0 | 1 | 0 |
| 0 | 0 | 1 | 0 | 0 |
| 0 | 1 | 0 | 1 | 0 |
| 1 | 0 | 0 | 0 | 1 |

Consider this as to be a character skeleton from which individual line segments have to identified[2]. Pixel in the top left corner is numbered as (1,1) and standard convention for numbering matrices is assumed so forth. Starters list in this image would be [(1,1),(1,5),(5,1),(5,5)]. Intersections would contain only the pixel (3,3). Now algorithm starts by processing the first starter i.e. (1,1). The next pixel would be (2,2) and the one after that would be (3,3). Pixel (3,3) is a intersection, so the algorithm stops the current segment, and declares all the neighbours as minor starters. So the minor starters list would contain [(4,2),(2,4),(4,4)]. The line segment formed now would contain [(1,1),(2,2),(3,3)]. Now the next starter ie (1,5) is considered. The next pixel is (2,4). Pixel (2,4) is a minor starter so the algorithm stops the current segment here and declares all unvisited neighbours of (2,4) , if it has any, as minor starters. Also (2,4) is removed from the minor starters list, since it has been visited. Now the next starter (5,1) is considered and the next line segment so formed would be [(5,1),(4,2)]. Pixel (4,2) would be removed from the starters list. In a similar fashion, the next line segment would be [(5,5),(4,4)]. So in end, there would be 4 line segments in total.

---

[2] It should be noted that this matrix is assumed to be obtained after zoning the original image.

## VIII. Distinguishing Line segments

After line segments have been extracted from the image, they have to be classified into any one of the following line types.
- Horizontal line.
- Vertical line.
- Right diagonal line.
- Left diagonal line.

For this ,a direction vector is extracted from each line segment which will help in determining each line type. For this, a convention is required to define the position of a neighbouring pixel with respect to the center pixel of the 3x3 matrix under consideration. The naming convention is as follows.

| 4 | 5 | 6 |
|---|---|---|
| 3 | C | 7 |
| 2 | 1 | 8 |

In the matrix given, 'C' represents the center pixel. The neighbouring pixels are numbered in a clockwise manner starting from pixel below the central pixel. To extract direction vector from a line segment, the algorithm travels through the entire pixels in the line segments in the order they forms the line segment. For instance, consider the line segment [(1,1),(2,2),(3,3),(4,4)]. The first pixel (1,1) is considered to be the current central pixel. With respect to this central pixel, (2,2) is in the position $8^3$. The first entry of direction vector would be 8. Next, pixel (2,2) is considered as central pixel. Now (3,3) is also in the position 8 with respect to this pixel. So the second entry of direction vector would also be 8. Going in this fashion, the direction vector for the given line segment would be (8,8,8). Though the above set of rules identifies all line segments, a drawback is that segments in the shape of 'V' or its rotated variations will be detected as a single line segment. For instance, in the given character image of 'A', the marked section is detected as a single line segment though it is composed of two entirely different line segments. To prevent such errors, a new set of rules is applied the segment given in the diagram will be detected as Once direction vector has been extracted, a new set of rules are applied to each direction vector to find new line segments inside it. The direction vector is subjected to following rules to find new linesegments.

[3]Refer the direction matrix

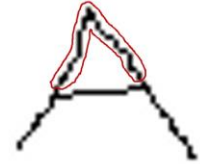

Fig. 4. Before applying direction rules

1) The previous direction was 6 or 2 AND the next direction is 8 or 4 OR.
2) The previous direction is 8 or 4 AND the next direction is 6 or 2 OR.
3) The direction of a line segment has been changed in more than three types of direction OR.

The line segment marked in the image was obtained before applying the direction rules explained last. Though this line segment is actually composed of two different line segments, it will be detected as one. But after applying the direction rules explained here, the two line types will be differentiated. If a new line segment is detected, then the direction vector is broken down into two different vectors at that point. Now the following rules are defined for classifying each direction vector.

1) If maximum occuring direction type is 2 or 6, then line type is right diagonal.
2) If maximum occuring direction type is 4 or 8, then line type is left diagonal.
3) If maximum occuring direction type is 1 or 5, then line type is vertical.
4) If maximum occuring direction type is 3 or 7, then line type is horizontal.

If two line types occur same number of times, then the direction type detected first among those two is considered to be the line type of the segment.

## IX. Feature Extraction

After the line type of each segment is determined, feature vector is formed based on this information. Every zone has a feature vector corresponding to it. Under the algorithm proposed, every zone has a feature vector with a length of 8. The contents of each zone feature vector are

1) Number of horizontal lines.
2) Number of vertical lines.
3) Number of Right diagonal lines.
4) Number of Left diagonal lines.
5) Normalized Length of all horizontal lines.
6) Normalized Length of all vertical lines.
7) Normalized Length of all right diagonal lines.
8) Normalized Length of all left diagonal lines.

9) Normalized Area of the Skeleton.

The number of any particular line type is normalized using the following method,

value = 1 - (( number of lines/10) x 2)

Normalized length of any particular line type is found using the following method,

length = (Total Pixels in that line type)/ (Total zone pixels)

The feature vector explained here is extracted individualy for each zone. So if there are N zones, there will be 9N elements in feature vector for each zone. For the system proposed, the original image was first zoned into 9 zones by dividing the image matrix. The features were then extracted for each zone. Again the original image was divided into 3 zones by dividing in the horizontal direction. Then features were extracted for each such zone.

After zonal feature extraction, certain features were extracted for the entire image based on the regional properties namely

- Euler Number: It is defined as the difference of Number of Objects and Number of holes in the image. For instance, a perfectly drawn 'A' would have euler number as zero, since number of objects is 1 and number of holes is 2, whereas 'B' would have euler number as -1, since it have two holes.
- Regional Area: It is defined as the ratio of the number of the pixels in the skeleton to the total number of pixels in the image.
- Eccentricity: It is defined as the eccentricity of the smallest ellipse that fits the skeleton of the image.

## X. Conclusion

This paper proposed a feature extraction technique that may be applied to classification of cursive characters for handwritten word recognition. The method proposed was tested after training a Neural Network with a database of 650 images. The algorithm was tested against a testing set of 130 images and 6 of them were detected erroneously.

In future, the method will be tried against standard databases like CEDAR. Also more experiments will be conducted with additional benchmark datasets.

## Acknowledgment

The author would like to thank Ms.Renu M.R., Department of Electronics, Amrita School of Engineering, for her valuable guidance and support for this work.

9) Normalized Area of the Skeleton.

The number of any particular line type is normalized using the following method,

value = 1 - (( number of lines/10) x 2)

Normalized length of any particular line type is found using the following method,

length = (Total Pixels in that line type)/ (Total zone pixels)

The feature vector explained here is extracted individualy for each zone. So if there are N zones, there will be 9N elements in feature vector for each zone. For the system proposed, the original image was first zoned into 9 zones by dividing the image matrix. The features were then extracted for each zone. Again the original image was divided into 3 zones by dividing in the horizontal direction. Then features were extracted for each such zone.

After zonal feature extraction, certain features were extracted for the entire image based on the regional properties namely

- Euler Number: It is defined as the difference of Number of Objects and Number of holes in the image. For instance, a perfectly drawn 'A' would have euler number as zero, since number of objects is 1 and number of holes is 2, whereas 'B' would have euler number as -1, since it have two holes.
- Regional Area: It is defined as the ratio of the number of the pixels in the skeleton to the total number of pixels in the image.
- Eccentricity: It is defined as the eccentricity of the smallest ellipse that fits the skeleton of the image.

## X. Conclusion

This paper proposed a feature extraction technique that may be applied to classification of cursive characters for handwritten word recognition. The method proposed was tested after training a Neural Network with a database of 650 images. The algorithm was tested against a testing set of 130 images and 6 of them were detected erroneously.

In future, the method will be tried against standard databases like CEDAR. Also more experiments will be conducted with additional benchmark datasets.

## Acknowledgment

The author would like to thank Ms.Renu M.R., Department of Electronics, Amrita School of Engineering, for her valuable guidance and support for this work.

## References

[1] M. Blumenstein, B. K. Verma and H. Basli, A Novel Feature Extraction Technique for the Recognition of Segmented Handwritten Characters, 7$^{th}$ International Conference on Document Analysis and Recognition (ICDAR '03) Eddinburgh, Scotland: pp.137-141, 2003.
[2] R. Gonzalez, E. Woods, Digital Image Processing, 2$^{nd}$ edition , Prentice hall.